\documentclass[runningheads]{llncs}
\usepackage{graphicx}
\usepackage{booktabs}
\usepackage[normalem]{ulem}
\useunder{\uline}{\ul}{}
\usepackage{longtable}
\usepackage{hyperref}

\begin{document}
\title{Federated Artificial Intelligence for Unified Credit Assessment}
\titlerunning{Federated Artificial Intelligence for Unified Credit Assessment}
% If the paper title is too long for the running head, you can set
% an abbreviated paper title here
%
\author{
Minh-Duc Hoang\inst{1} \and
Linh Le\inst{2} \and
Anh-Tuan Nguyen\inst{2} \and
Trang Le\inst{2} \and \\
Hoang D. Nguyen\inst{3}}
\authorrunning{Hoang et al.}
% First names are abbreviated in the running head.
% If there are more than two authors, 'et al.' is used.
%
\institute{University of New South Wales Sydney, Australia \\
\email{MinhDuc.Hoang@unsw.edu.au} \and
BeU Research Group, Vietnam \\
\email{\{Linh.Le, Tuan.Nguyen, Trang.Le\}@beu.ai} \and
University of Glasgow, Singapore\\
\email{HarryNguyen@glasgow.ac.uk}}
\maketitle              % typeset the header of the contribution
\begin{abstract}

With the rapid adoption of Internet technologies, digital footprints have become ubiquitous and versatile to revolutionise the financial industry in digital transformation. This paper takes initiatives to investigate a new paradigm of the unified credit assessment with the use of federated artificial intelligence. We conceptualised digital human representation which consists of social, contextual, financial and technological dimensions to assess the commercial creditworthiness and social reputation of both banked and unbanked individuals. A federated artificial intelligence platform is proposed with a comprehensive set of system design for efficient and effective credit scoring. The study considerably contributes to the cumulative development of financial intelligence and social computing. It also provides a number of implications for academic bodies, practitioners, and developers of financial technologies.

\end{abstract}
\section{Introduction}
The recent decades have witnessed a rapid pace of change with the advance of the Internet and new technologies. Having all these digital technologies available at their fingertips, people are using them in everyday life, making them ubiquitous \cite{wang2007social}. The financial industry, therefore, has been undergoing digital transformation; in which new types of data, products and services have been continuously developed \cite{scardovi2017digital}. 

To date, there are approximately 1.7 billion adults that are unbanked according to The Global Findex database \cite{demirguc2018global}, especially those who are living in developing countries where transactions are predominantly facilitated in cash. It is significantly challenging for financial institutions to evaluate credit inquiries or loan applications; nevertheless, digital footprints such as social network or mobile data have become increasingly promising to address this challenge in digital transformation. These new types of data can be used to cover this deficiency of available data in financial risks assessment with broader coverage of information \cite{masyutin2015credit}. The past studies have shown that behavioural insights on digital data are a good supplement to the traditional credit scoring approach as it can help banks to predict fraud, reduce defaults and avoid leaving out potential credit inquiry \cite{bjorkegren2018potential,ge2017predicting,tan2018social}.

Digital footprints involve dealing and collecting enormous amounts of data from a great variety of sources in different types and semantics, creating a significant challenge to unify, aggregate, classify and fulfil all data fields in one system. Moreover, they are often associated with issues of data reliability, especially in the measure of soft factors and risks of complex fraudulent data. The use of artificial intelligence (AI), hence, plays a vital role to harness big data for better financial products and services. This paper aims to address major boundaries of credit scoring on fiscal and social footprints by introducing a federated credit assessment framework with a comprehensive set of system concepts and design guidelines.

Based on theoretical foundations, the research contributes to the cumulative development of financial artificial intelligence and risk scoring systems. It draws out many implications for academic bodies, industrial practitioners, and developers.
 
The structure of the paper is as follows. Firstly, we review the literature background of our paper in Sect. 2. Secondly, we discuss the existing challenges of the credit assessment in Sect. 3. Next, we present our federated credit assessment framework with system design and concepts. Lastly, Sect. 5 concludes our paper with findings and contributions.

\section{Literature Background}
\subsection{Financial Creditworthiness Scoring}

\begin{table}[htb]
\caption{Common Financial Predictors}
\label{tab:fin}
\centering
\setlength{\tabcolsep}{0.2cm}
\begin{tabular}{p{1.8cm}p{3.2cm}p{4.5cm}p{1cm}}
\hline
Category &
  Description &
  Example Indicators &
  Ref \\ \hline
Demographic &
  Personal details that matter to the assessment of an individual’s creditworthiness &
  \textbf{Numerical:} age, number of children, number of dependent, monthly income;  \textbf{Categorical:} gender, marital Status, residential status, postcode, home/mobile phone, education, profession &
  \cite{west2000neural,dinh2007credit} \\
Tenure &
  Variables that are related to time &
  Time at current address, time at present job, time in employment, time with the bank, account longevity, duration of loan in months &
  \cite{desai1996comparison,sinha2004evaluating,malhotra2003evaluating} \\
Trade &
  Variables related to an individual’s trading activities &
  \textbf{Numerical:}  number of active accounts, number of open accounts, percent trades that delinquent/never delinquent, ratio of payment/total income, ratio of debt/total income; \textbf{Categorical:}  type of bank accounts, loan from other banks &
  \cite{desai1996comparison,sinha2004evaluating} \\
Balance &
  Variables that help access the financial situation of an individual &
  Past and current amount of loans received, saving account status, current account status &
  \cite{dinh2007credit,sinha2004evaluating} \\
Inquiry &
  These variables illustrate the underlying factors related to every credit application &
  \textbf{Numerical:}  number of inquiries in the past months, credit limit; \textbf{Categorical:} inquiry purpose, loan guarantors, collateral type, credit rating &
  \cite{desai1996comparison,sinha2004evaluating,lee2005two} \\ \hline
\end{tabular}
\end{table}

Credit scoring has been defined as a set of models that supports the process of financial decision making in regard to the granting of credit for individuals by lenders \cite{thomas2017credit}. Since the 1950s, the concept and techniques of credit assessment have evolved swiftly, supported by the development of expertise and technology to match the increase in customer demand after the advent of credit card \cite{mays2001handbook}. Advanced technologies such as data mining, machine learning and AI have been providing tremendous support to the creation of new models that ensure better accuracy to predict individual’s risk performance. Furthermore, the application of credit scoring has also transformed beyond its initial use in the lending industry to predictive analytics on how likely a customer will churn, which patients might be posing a higher risk of a specific kind of disease or whose tax submission should be investigated further \cite{thomas2017credit}. 

Empirical studies have shown a great variety of models being applied to generate credit scores for bank users throughout the last several decades \cite{chen2018interpretable,mays2001handbook,lee2005two}. In general, there is a certain level of similarity between these models in terms of the predictors being used to produce the response. However, different models will have their own selection for the number of predictors used, and there are also some distinct predictors that are considered to add novel values to the existing models. \textbf{\autoref{tab:fin}} lists out some of the most commonly used predictors in popular financial credit scoring models. 

Nevertheless, with the increasing ubiquity of digital technologies, the finance industry has been facing unprecedented challenges in assessing credit risk. Hence, there is an emerging trend for application of social data into many aspects of credit scoring.

\subsection{Social Reputation Scoring}
With the average worldwide Internet penetration of 59\% today \cite{statista2020global}, digital footprints are being created by billions of monthly active users, which contribute to the pool of an enormous amount of social data. Hence, combining social and financial data into the assessment of credit risks has become a new trend with great potentials of social data in the lending industry. Social data serve as a valuable and supplementary data source to enhance predictive models in credits.

A large number of studies have extensively investigated the use of digital footprints for credit scoring \cite{guo2016footprint,ge2017predicting,tan2018social,masyutin2015credit}.  Social footprints include user demographics (e.g., age, gender, relationship status), user-generated contents (e.g., Twitter tweets or Facebook posts), and social relationships (e.g., friends or followers) on social networking sites. Guo et al. (2016) demonstrated significant improvement by 17\% with the use of social features in their personal credit scoring model \cite{guo2016footprint}. In addition, mobile data (e.g., call and messaging events) and browsing behaviours (e.g., websites and browsing duration) have also been explored to improve credit models \cite{oskarsdottir2019value}. San et al. (2015) employed Gradient Boosting Tree in their real-time approach using mobile behaviours, which outperformed credit bureaus, such as Experian or Equifax,  by a large margin \cite{san2015mobiscore}. With a huge amount of unstructured data, these social data sources provide capabilities to track and evaluate individuals and businesses for trustworthiness, not only in a financial credit sense but also in a behavioural sense, i.e. sincerity, honesty, and integrity. 

As human interactions are now taking place more and more often on the Internet, social data have not only been utilised by private sectors in sharing economy and collaborative economy, but also by public sectors in social credit scoring. Social credit has become increasingly beneficial in the identification and dissemination of reputational information, which has become a crucial element of the 21st-century digital transformation. With the recent advancement of big data and artificial intelligence technologies, social-based credit systems have evolved beyond traditional credit scoring approaches into massive and complex modelling systems with the utilisation of technological data (e.g., facial recognition or movement trackings).

\section{Emerging Challenges in Credit Assessment}

With the advents of new technologies and the development of expertise across the industry, the traditional credit scoring methods have now transformed to become automated as banks started looking for a way to utilise their limited resources to no longer have to "treat each small exposure individually" \cite{vojtek2006credit}. However, along this much-needed digital transformation are concerns about different obstacles that can challenge the implication of these new approaches. Hereby, this paper will look into several challenges that current credit assessment systems are facing as the following.  

\begin{itemize}
    
    \item \textbf{Big and heterogeneous data}. It is without a doubt that big data is playing an imperative role in improving the accuracy of credit scoring given the values it can create by providing accesses to a larger pool of information over a broader target audience. However, big data is also a double-edged sword that poses a variety of challenges due to its complex and dynamic characteristics. Collecting this enormous amount of data requires the contribution from a great variety of sources / websites / applications in different languages and formats, targeting various segmentation of users. Thus, it creates a significant challenge to unify, aggregate, classify and fulfil all data fields in a scoring system \cite{wang2017heterogeneous}. Even though this can be partially overcome by certain methods, it is laborious to choose proper objective functions for a specific data heterogeneity problem \cite{mandreoli2019dealing}. Moreover, existing approaches might overlook the actual context and interpretation of the data, thereby compromising the performance of the system with a significant impact on the targeted users. \\ 
    
    \item \textbf{Reliability}. The social-financial credit scoring has commonly been criticised for issues of data reliability, particularly in the measurement of soft factors and risks of a monopolistic system lacking cross-referencing. It is possible for individuals or platform providers might deliberately manipulate data resulting in inaccurate social-based credit ratings. Or simply, another issue regarding data reliability lies on human errors of the input data. Working with a model that heavily relies on the quality of the input, especially historical data in this case, requires proper data validation techniques. Moreover, getting up-to-date data plays a crucial role to improve system performance \cite{thomas2017credit}. \\
    
    \item \textbf{Security and Privacy}. Credit scoring also raises controversy for its potential infringement on privacy. Currently, data protection law through Council of Europe convention and EU Directive, and most recently General Data Protection Regulation in EU (GDPR), which introduce a framework for the regulation of rating and reputation data, has been growing in significance. Data protection laws also elaborate further on the conditions of legal processing of personal data. Nevertheless, from a policy perspective, it is suggested that a review of security and privacy should be in place as a means of safeguarding individual rights as well as exercising checks and balances of the design and operation of the credit assessment systems. It is important to highlight any problems of information security, data privacy and flaws, particularly with respect to ratings by financial institutions. \\
    
    \item \textbf{Explainability}. A credit assessment system will not be trusted without transparent. Interpretability was the key selling point in traditional credit models; nevertheless, AI-based models with new types of data have been tagged with a lower degree of explainability. This has become a technological challenge of explaining AI decisions. Credit ratings of individuals based on social and financial data are prone to errors due to data, algorithms, or manipulation issues. Therefore, explainable and accountable credit assessment is necessary to build and engage sincerity and honesty in consumer networks. \\
    
\end{itemize}

\section{Federated Artificial Intelligence for Unified Credit Assessment}

\begin{figure}[t]
\includegraphics[width=\textwidth]{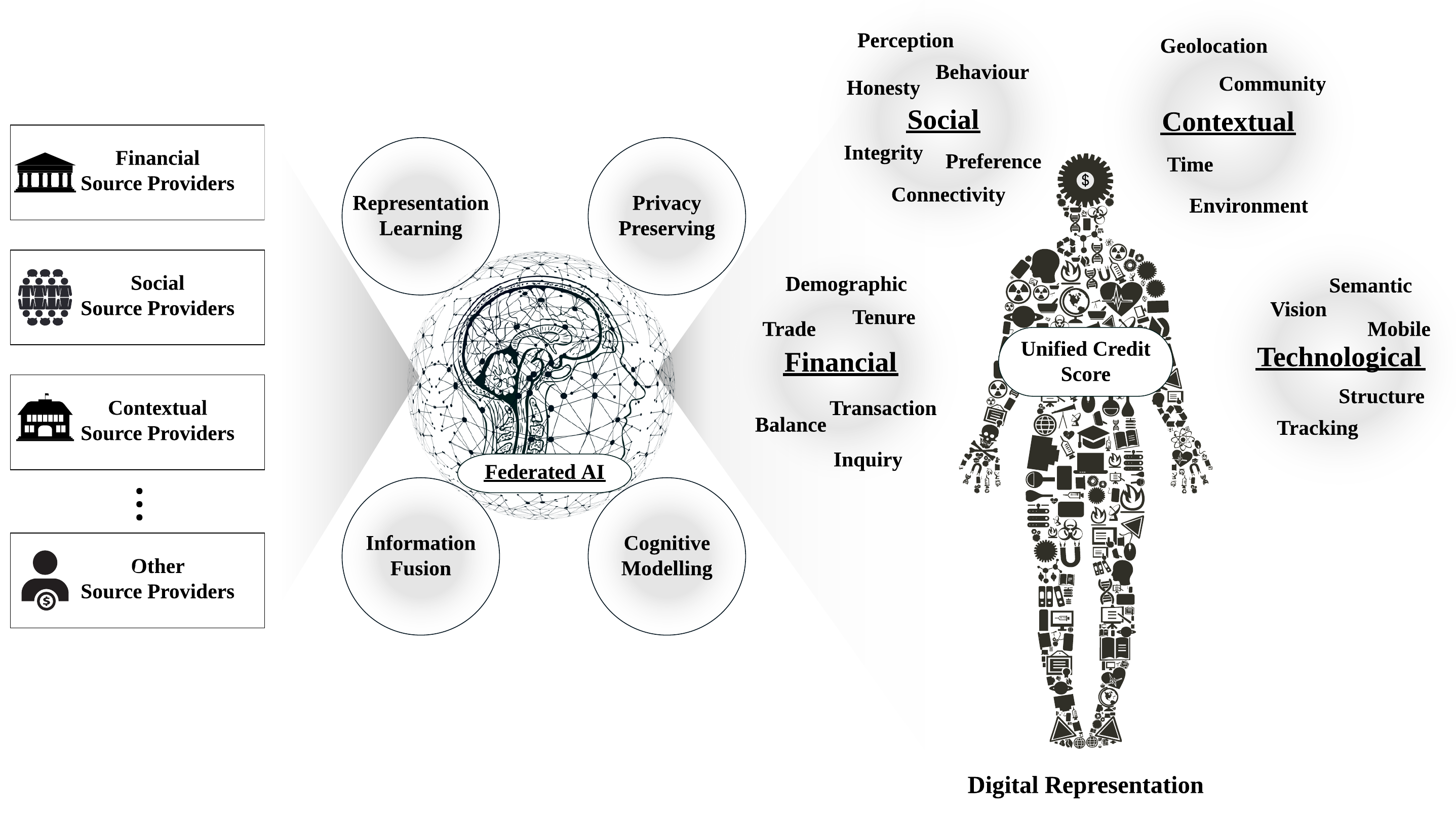}
\caption{Federated Artificial Intelligence for Credit Assessment} \label{fig:overview}
\end{figure}

In this paper, we propose a federated credit assessment framework with the use of artificial intelligence to address the existing challenges in credit scoring. It utilises a state-of-the-art machine learning engine to integrate heterogeneous data sources and to produce high-dimensional digital representations of consumers. The dimensionality of these digital representations may range from 20,000 to 30,000 data features; which encompass social, financial, contextual, and technological characteristics optimised under millions of artificial neurons according to the criterion based on credit decisions.  \textbf{Fig. \ref{fig:overview}.} provides an overview of the federated artificial intelligence for unified credit assessment.

The framework consists of FIVE (5) key components: (1) Unified credit score, (2) Information fusion, (3) Privacy-preserving, (4) Cognitive Modelling, and (5) Representation learning. The following sub-sections discuss these components in great details.

\subsection{Unified Credit Score}

\begin{table}[htb]
\caption{Unified Credit Score - Types and Dimensions}
\label{tab:ucs}
\setlength{\tabcolsep}{0.2cm}
\begin{tabular}{p{1.75cm}p{1.75cm}p{7.5cm}}
\hline
Type &
  Dimension &
  Description \\ \hline
Financial &
  Demographic &
  Personal details, including consumer profile, backgrounds, and socioeconomic measures \\
 &
  Transaction &
  Transactional records, including purchases, inquiries and transfers \\
 &
  Credit &
  Behavioural features related to historical credit activities \\
 &
  Tenure &
  Time analysis features related to banking activities \\ \hline
Social &
  Behaviour &
  Social features related to how an individual conducts oneself in social networking sites  \\
 &
  Preference &
  Personal preferences and habits based on social profile \\
 &
  Perception &
  Data features related to emotions, honesty, integrity, etc \\
 &
  Connectivity & 
  Social relationships including social networks associated with credit ratings and communications \\
 &
  Content &
  Unstructured data features generated by consumers in social networking sites \\ \hline
Contextual &
  Geolocation & 
  Geographical features, including location-based conditions and networks \\
 &
  Time & 
  Temporal and seasonal features \\
 &
  Environment & 
  Environmental features, including epidemiological and pollution conditions  \\
 &
  Community & 
  Physical context features, including neighbourhood, groups and businesses \\ \hline
Technological &
  Semantic &
  Linguistic and philosophical features based on natural language modelling \\
 &
  Vision &
  Visual features, including facial features and user-generated images/videos \\
 &
  Mobile &
  Activity data features based on mobile usage and browsing behaviours \\
 &
  Tracking &
  Movement-based features including sensor-based movements and positioning measures \\ \hline
\end{tabular}
\end{table}

Credit score is traditionally assessed based on hard data denoting to contextual and financial information of borrowers such as age, name, address, credit history, and transactions. Nevertheless, due to the advent of social media and technology advancement, personal credit score has been evaluated using multiple categories of data, including hard and soft information \cite{guo2016footprint,san2015mobiscore}. This research paper proposes a unified credit score which is assessed based on wider sets of data. We suggest the construction of a credit score based on four types with multiple dimensions: financial, social, contextual and technological characteristics, as shown in \textbf{\autoref{tab:ucs}}.

It aims to provide integrated and comprehensive aspects of the digital representation for each individual. Both hard data (financial and contextual data) and soft data (social and technological information) are included in the computation of the unified credit score. For instance, the financial data include but not limited to credit records (e.g., credit score, debt to income ratio and annual income) and transactions (e.g., number of successful sale, days from last purchase, and transaction ratio). In the social category, a useful credit model needs to investigate behaviours such as trends in commercial transactions or social network characteristics in online platforms. Moreover, content-related features, including numbers of friends, posts, messages, or interactions, are good supplement to enhance credit scoring. Also, contextual dimensions such as geolocation, time, environment, and community play a crucial role in predicting the likelihood of default behaviours. Semantic, visual, mobile and tracking features are promising predictors of the technological dimensions, which enrich other types of data sources for better interpretability and performance.

\subsection{Information Fusion}

Unified credit assessment entails individual-level data analysis from multiple, heterogeneous data sources. An enormous amount of financial, social, contextual and technological data are collected from a vast variety of sources / websites / applications. They are diverse in languages, types, formats, and even unit levels; thus, it is essential to employ information fusion strategies to eliminate uncertainty and reliability issues in the big data. The process of information fusion integrates multiple data sources into a robust, accurate and consistent body to be used in the federated AI. This component involves many data processing techniques such as aggregation, selection, cleaning, construction, and formatting. For instance, a hierarchical decomposing method can be adopted to handle the data at multiple levels, such as individual and contextual categories. Moreover, we introduce the use of multilayer network (MLN) analysis to model social data from multiple perspectives for higher efficiency \cite{vu2019generic}.

Furthermore, digital footprints of consumers are typically widespread on multiple financial information systems and social networking sites. These data traces are found and collected in a disconnected and fragmented manner. User entity resolution, therefore, plays a crucial role to interconnect the data traces into single digital identities. Transfer models for cross-domain user matching are proposed to incorporate consumer activity data into user entity resolution with deep neural networks \cite{ahangama2019application}. This technique outperforms similarity-based matching models to provide a mechanism for recognising and merging consumer profiles. We propose the use of deep user activity transfer to prevent duplicated profiles and synthetic data, thereby increasing the reliability of the data.

\subsection{Federated Representation Learning}

\begin{figure}[htb]
\includegraphics[width=\textwidth]{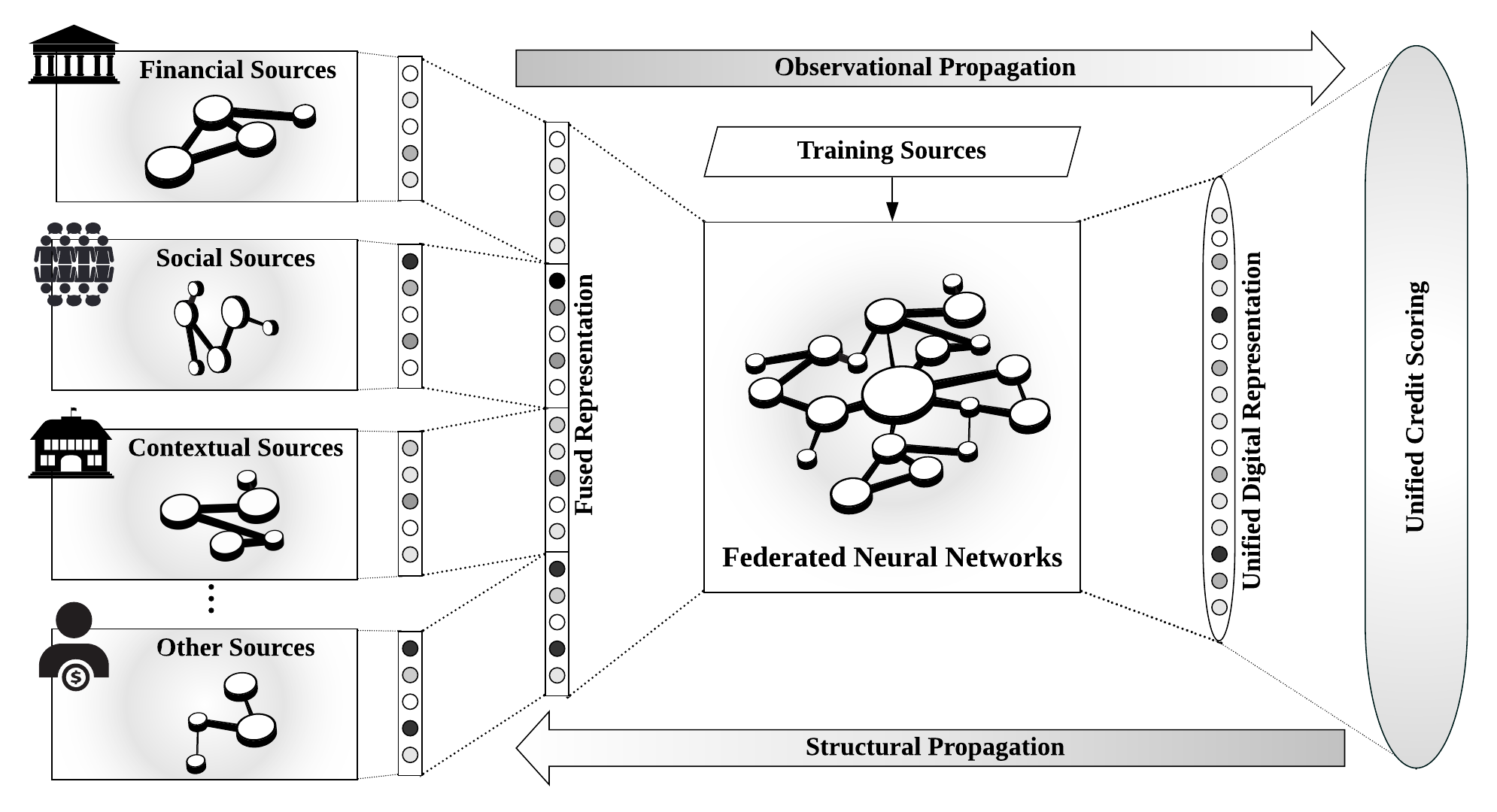}
\caption{Federated Representation Learning} \label{fig:fdl}
\end{figure}

This paper introduces federated representation learning that builds deep neural network models across decentralised data providers. This approach enables multiple organisations to optimise the learning pathways with a unified objective without sharing data. The heterogeneity and security issues of data sources, therefore, are taken care of in the federated AI. \textbf{Fig. \ref{fig:fdl}.} illustrates how federated representation learning works across multiple data source providers.

The federated learning aims to orchestrate data representations from various sources into a unified digital body for credit scoring in an iterative manner. Each organisational unit has dynamic and local neural networks with a standardised objective function to observe incoming data samples for producing high-dimensional representations, which typically range between 2,048 and 4,096 data features. This approach harnesses parallelised computing powers to learn distributed and heterogenous data effectively. Moreover, the use of artificial neural networks allows federated AI to handle complex and non-linear data structures.

The main principle of federated representation learning consists in the dynamic nature of neural networks with constant observational and structural propagation. On the one hand, observational propagation is a forward process that represents and computes data observations through a system of millions of neurons in order to elicit the loss in credit assessment. On the other hand, structural propagation happens with the training of labelled data sources to provide the exchange of neural network weights and reconfigurations between a central node and local nodes. As a result, multiple data sources from various organisations can be efficiently integrated in harmony to score personal credits.

\subsection{Privacy-Preserving}

The key advantage of federated AI is to ensure data secrecy, in which it is by design that no local data is transmitted externally for machine learning. Nevertheless, there is a possibility for individual re-identification based on high-dimensional feature matching. With the use of publicly available information, a reverse-engineering process can be done to discover consumer profiles. 

We propose the use of user-level differentially private representations to balance the trade-off between privacy and data utility \cite{vu2019dpugc}. Our federated AI employs state-of-the-art neural network architecture to capture the sensitivities of consumer data and to decide the degree of representation learning while achieves good performance for credit scoring.

\subsection{Cognitive Modelling}

\begin{figure}[htb]
\includegraphics[width=\textwidth]{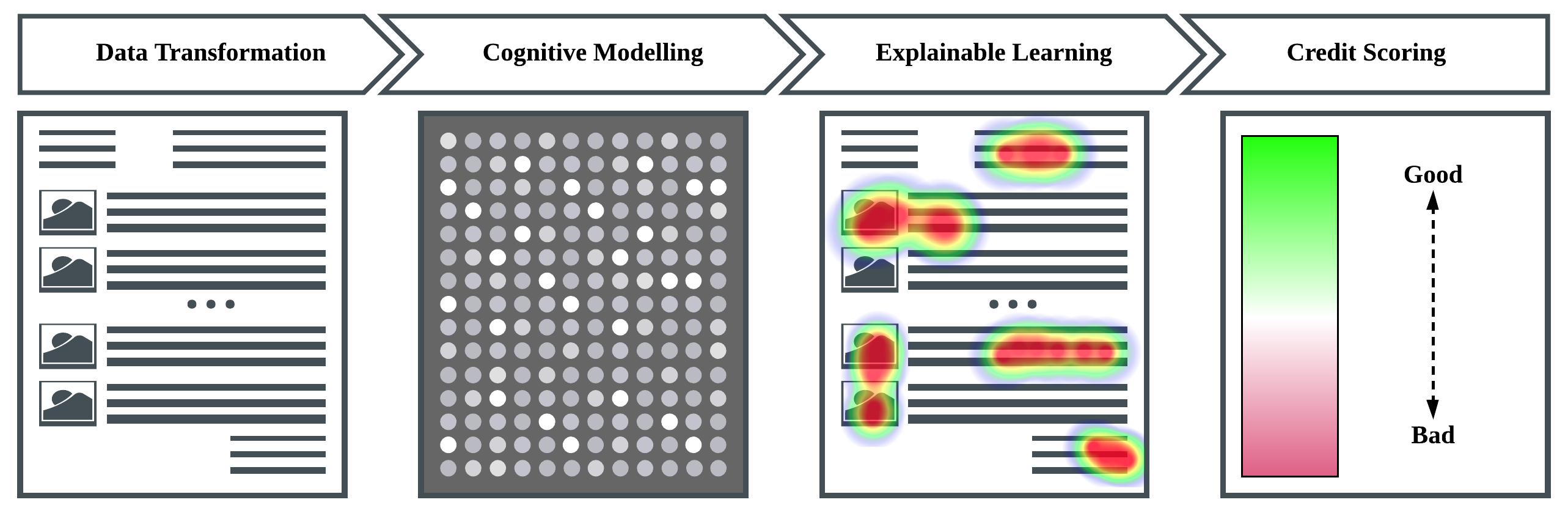}
\caption{Cognitive Modelling for Credit Assessment} \label{fig:ci}
\end{figure}

In recent years, cognitive intelligence has been widely explored as an advancement of brain-inspired systems in artificial intelligence \cite{wang2016cognitive}. Although the mechanism of human brains remains largely unknown, simulating cerebral activities creates new ways of understanding data and making decisions, especially for credit scoring. In this study, the ultimate objective is to develop a brain-inspired artificial intelligence to assess personal credits similar to human-based decisions. \textbf{Fig. \ref{fig:ci}.} provides the general concept of cognitive modelling for credit assessment.

We propose several steps to develop our explainable, federated AI. First, transforming different types and formats of data into relevant data embeddings. Second, projecting data embeddings into user-level cognitive maps with multiple distributed data features. Cognitive modelling happens at a deeper level of computational consciousness in alignment with human judgements in investigating personal credits. Furthermore, explainable learning is introduced to provide stakeholders of the cognitive intelligence system an interpretation of the predictions made by the deep learning model using the state-of-the-art approaches.

\subsection{Discussion}
Unified credit assessment is promising to provide integrated and comprehensive representation for each individual. With the support of federated AI, the proposed approach can be applied in determining individual creditworthiness for many applications, including property loan or personal lending. 

Online B2P (business-to-peer), P2B (peer-to-business), and P2P (peer-to-peer) lending business models have become increasingly popular worldwide. These operations contribute to the considerable market growth with huge demand and supply; where banks are unwilling to lend to individuals due to the lack of an established system for assessing consumer credit risk. On the supply side, many households have reasonable savings sitting in their bank accounts with low-interest rates, and there limited options for avers to manage their money. As a result, the consequent demand for investment options is leading to online personal lending. A problem of personal lending lies on the difficulty to assess financial creditworthiness and degree of trustworthiness on the data used in the analysis. The problem persists more burdensome in developing countries; in which the majority of people do not use bank accounts, leaving social traces even more critical in assessing individual creditworthiness.

The proposed framework will help B2P, P2B, P2P lending providers easier in making credit decisions. It is essential for lenders to identify creditworthiness by investigating the digital representations of their potential customers with the use of financial and social data. Unified credit scoring, therefore, will be a useful tool to improve both customer ratings and transaction monitoring. Moreover, it has the potential to improve the effectiveness of anti-money laundering programs; which meaningfully contributes to the economic and social stability.

\section{Conclusion}
Our study has several implications for theoretical literature and development of financial technologies. First, we propose a new paradigm of unified credit assessment supported by federated artificial intelligence. Second, we elaborate this paradigm to recommend a novel distributed system to produce digital representations for individual credit scoring with the privacy-preserving mechanism. Last but not least, we designed the federated AI platform, which is capable of reshaping current credit assessment approaches towards data and decision orchestration. It unveils the capability of building a unified credit score based on financial, social, contextual, and technological data for effective credit evaluation.

This paper is not an end, but rather a beginning of future research. We are looking into ways of further refining machine learning through the process of state-of-the-art brain-inspired technologies. 

\section{Acknowledgment}
This study is a part of the BeU AI research initiative.

%
% ---- Bibliography ----
%
% BibTeX users should specify bibliography style 'splncs04'.
% References will then be sorted and formatted in the correct style.
%
\bibliographystyle{splncs04}
\bibliography{reference}
\end{document}